\pdfoutput=1

\documentclass[11pt]{article}

\usepackage[]{ACL2023}

\usepackage{times}
\usepackage{latexsym}

\usepackage[T1]{fontenc}

\usepackage[utf8]{inputenc}

\usepackage{microtype}
\usepackage{inconsolata}

\usepackage[utf8]{inputenc} 
\usepackage[T1]{fontenc}    
\usepackage{hyperref}       
\usepackage{url}            
\usepackage{booktabs}       
\usepackage{amsfonts}       
\usepackage{amsmath}
\usepackage{nicefrac}       
\usepackage{microtype}      
\usepackage[pdftex]{graphicx}
\newcommand{\R}{\mathbb{R}}

\title{Feature-augmented Machine Reading Comprehension with Auxiliary Tasks}
 \author{Yifeng Xie \\
 School of Mathematics and Statistics,\\
Guangdong University of Technology, China}

\begin{document}
\maketitle
\begin{abstract}
While most successful approaches for machine reading comprehension rely on single training objective, it is assumed that the encoder layer can learn great representation through the loss function we define in the predict layer, which is cross entropy in most of time, in the case that we first use neural networks to encode the question and paragraph, then directly fuse the encoding result of them.
However, due to the distantly loss back-propagating in reading comprehension, the encoder layer cannot learn effectively and be directly supervised. Thus, the encoder layer can not learn the representation well at any time. 
Base on this, we propose to inject multi granularity information to the encoding layer.
Experiments demonstrate the effect of adding multi granularity information to the encoding layer can boost the performance of machine reading comprehension system. Finally, empirical study shows that our approach can be applied to many existing MRC models.
\end{abstract}

\section{Introduction}
\begin{figure*}
  \centering
  \includegraphics[scale=0.5]{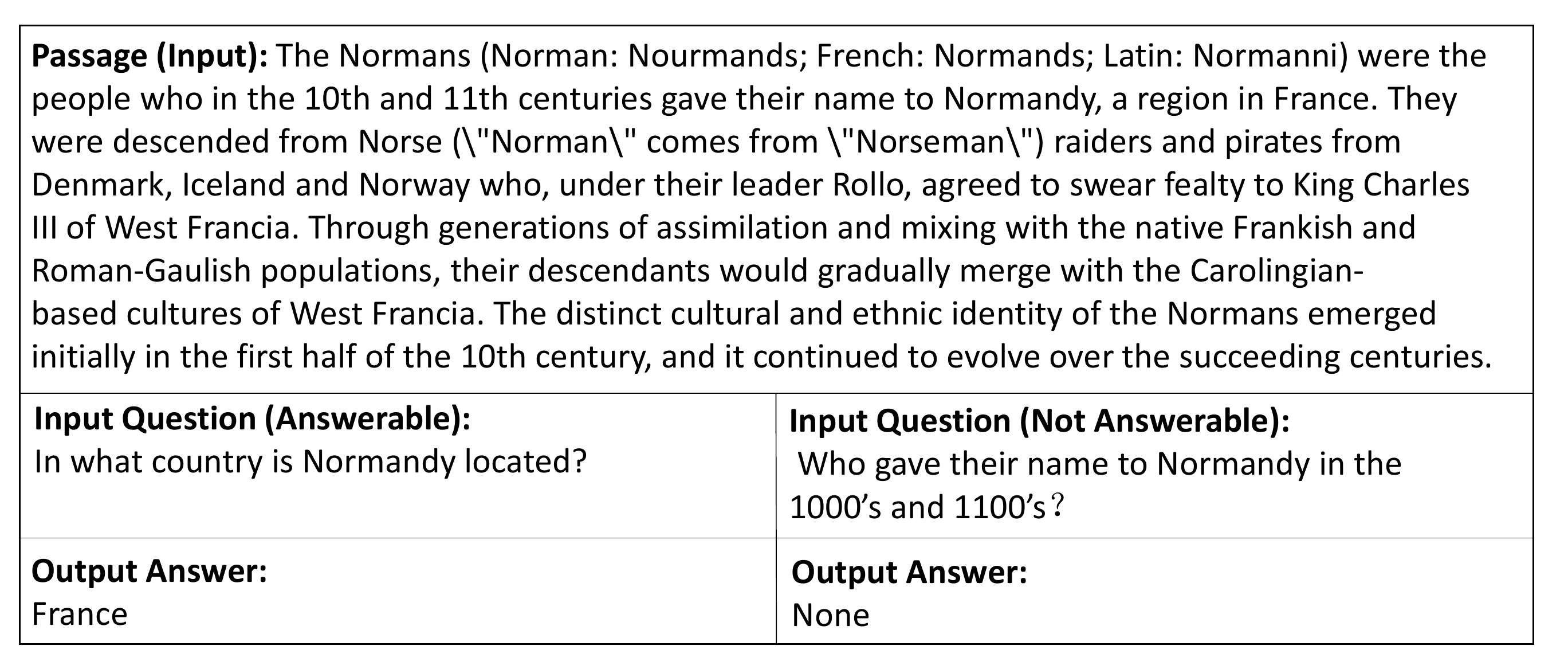}
  \caption{An example of SQuAD2.0.}
  \label{fig:squad}
\end{figure*}

Machine Reading Comprehension (MRC) is a field having gained tremendous popularity among researchers in the last few years. In this field, a MRC model is designed to process and understand the context and the question in order to provide a reasonable answer. 
Recently,~\citet{Rajpurkar2018squad2.0} released the SQuAD 2.0 dataset, which, compared to SQuAD 1.1~\cite{rajpurkar2016squad}, incorporates unanswerable questions to make it more difficult to answer the questions accurately.
Figure~\ref{fig:squad} gives an example of the MRC task, whose inputs are passage and question, and predict the answer as output to the question.
A large number of models have been explored for this question answering task, including RNN-based models, CNN-based models, transformer-based models, and pre-trained language models, and have been shown to be useful on SQuAD 2.0.

\begin{figure*}[t]
  \centering
  \includegraphics[scale=0.48]{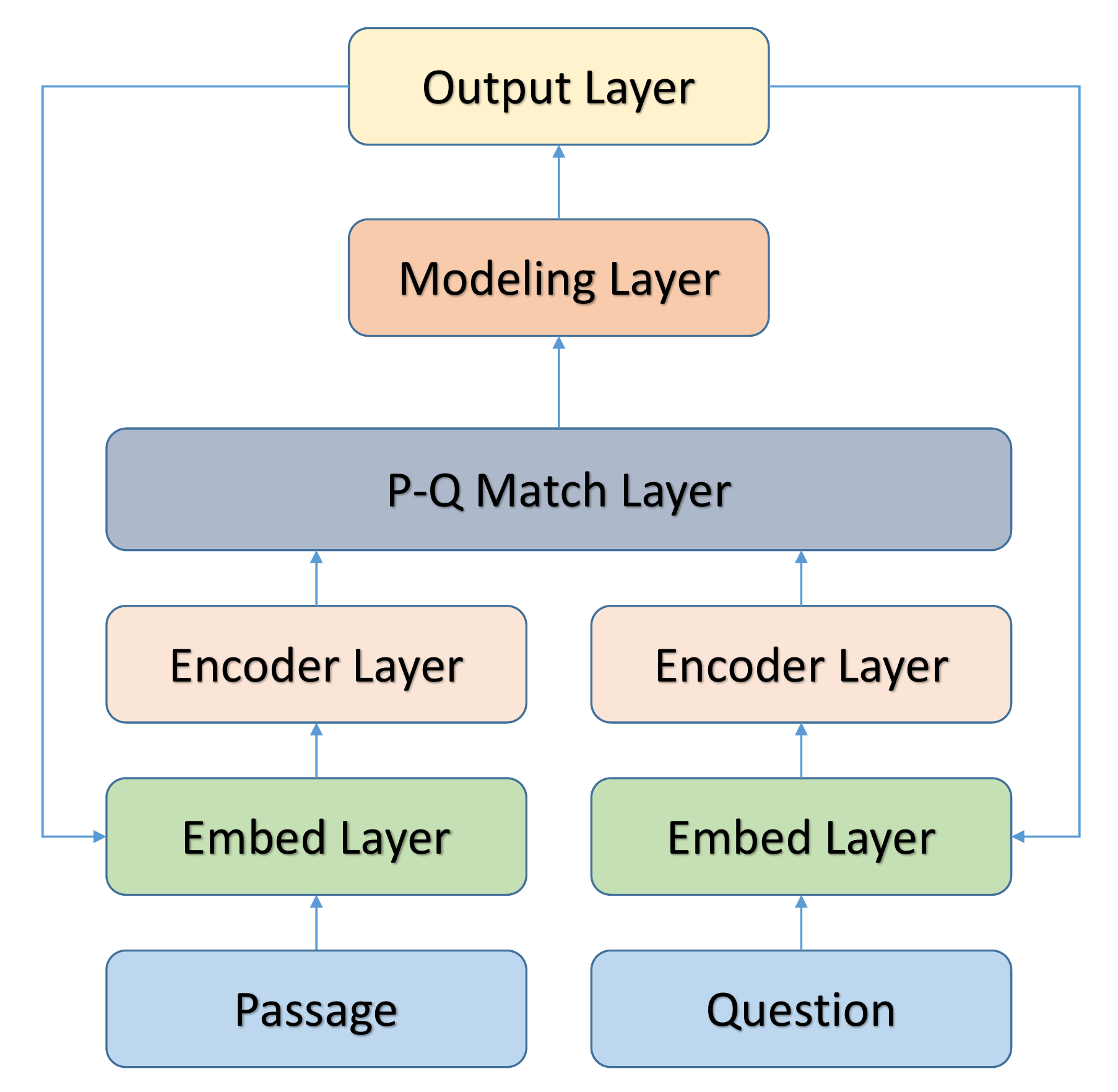}
  \caption{Conventional Machine Reading Comprehension System.}
  \label{fig:label}
\end{figure*}

Conventional MRC framework can be seen in~\ref{fig:label}, where passage and question are encoded seperately and then passed to a match layer for chossing the best answer candidate.
However, due to the distantly loss back-propagating in MRC, the encoder layer need to experience a long distant parameter update and could not be learnt effectively without directly supervised, which may lead to the bad representation ability of encoder layer.
As we all know, encode the text sequence with better representation plays a vital role in various NLP task.
What's more, this becomes more obvious when encountering the MRC task, because of the complex and multi encoding layer most MRC system used~\cite{Seo2017BiDAF,Wang2017MatchLSTM,Wang2017R-Net,Xiong2017dynamic,chen2021adaptive}.

Ideally, the encoding layer should be able to learn multi-granular text information. 
Inspired by the great performance of jointly training in Spoken Language Understanding (SLU)~\cite{Qin2019Stack-Propagation,chen2022joint,xu2021semantic,zhou2022calibrate,Zhu2022graph,zhou2020pin,huang2020FLSLU,huang2022slt,huang2021sentiment,chen2022han,chen2022bilinear}, of which Intent classification (IC) and slot filling (SF) are two main tasks, we know that fine-grained and coarse-grained tasks can effectively complement and boost each other.
Based on this, we propose to use directly loss back propagation to strengthen the encoding with both coarse-grained and fine-grained information.

This work describes a stack learning framework for Machine Reading Comprehension, which aims to strengthen the encoder layer by introducing multi-granularity text task.Specifically, we use IC task and SF task to extract the multi-granularity information of question and passage, then inject them to the encoder layer of MRC system. 
Intent classification module can encode the sentences with coarse-granularity language embedding, which can better learn the representations at sentence level.
Slot filling module can encode the sentences with fine-granularity language representation, to better capture the respective representations at token level.
Then, the encoding layer of SLU is used to fully fuse information from both sentence level and token level representations in the MRC module.
In this paper, we adopt a classical Bi-DAF model for both single passage and multi passage machine reading comprehension as a baseline. Jointly train the paragraph selection and paragraph span extraction model for reduce the distantly loss back-propagating problem, in order to further boost the performance of Machine Reading Comprehension.

Overall, this paper proposes to extract multi-granularity representations as the external input features for the basic MRC model. The contributions of our paper are as follows:

\begin{itemize}
    \item We propose to jointly learn coarse and fine text information to strengthen the encoder for Machine Reading Comprehension.
    \item Experiment on three model shows that our propose framework have obvious improved, while use strong language model, i.e. BERT, as the baseline, our model get remarkable results in SQuAD2.0.
\end{itemize}

\section{Related Work}
\subsection{Machine Reading Comprehension}
Reading comprehension, which aims to answer questions about a document, has become a major focus of NLP research. Many algorithm has been proposed to solved this problem. A model must be able to process and understand a context and question in order to provide a reasonable answer. 

A large amount of model has been explored to this question answering task, including RNN-based model, transformer-based model, Pre-trained Contextual Embeddings model, Pre-trained Language Module, and have been proved to be effective on this dataset~\cite{Seo2017BiDAF,Wang2017MatchLSTM,Xiong2017dynamic,kadlec2016text,sordoni2016iterative,Devlin2019BERT,yang2019xlnet,hou2020dynabert}.\nocite{huang2021ghostbert}

\subsection{Spoken Language Understanding}
Many models have been proposed to solve the intent classification and slot filling problems, and they are the two main tasks of SLU~\cite{chen2022han,chen2022bilinear,chen2022joint,huang2021sentiment,huang2020FLSLU,Zhu2022graph,zhou2020pin}.
Depending on whether intent classification and slot filling are modeled separately or jointly, past models have been classified into independent modeling approaches~\cite{zhou2022calibrate,xu2021semantic} and joint modeling approaches~\cite{Chen2019jointbert}.

\nocite{zhou2022equivariant,xie2022decoupled}

\subsection{Feature Integration}
Inspired by the successful usage of Stack-Propagation Framework in Spoken Language Understanding \cite{Qin2019Stack-Propagation} and successful usage of implicit representation integration in Semantic Role Labeling \cite{xia2019SRL}, 
as well as feature fusion and enhancement methods across different modalities and tasks~\cite{chong2022masked,zeng2022low,zhou2022mets,huang2021audioMRC,huang2021audioMRC2},
we design a jointly learning framework to train MRC and SLU tasks interactively, so as to reduce the distantly loss backward problem.

\section{Conventional Text Feature Representation in MRC}

\subsection{Input Representation}
Following \cite{xia2019SRL}, we utilize CNNs to encode characters for each word $\vec{E}_{i}$ into its character representation, denoted as $\vec{E}_{i}^{\text {char }}$. Then, we employ word embedding Glove to represent the word-level features, denoted as $\vec{E}_{i}^{\text {word }}$. Besides, we employ BERT feature representations \cite{Devlin2019BERT} to bring more representation message in our model, which we denote as $\vec{E}_{i}^{\text {BERT }}$. 
Formally, the input representation of $\vec{E}_{i}$ is:

\begin{equation}
\vec{E}_{i}=\vec{E}_{i}^{\text {char }} \oplus \vec{E}_{i}^{\text {word }} \oplus \vec{E}_{i}^{\text {BERT }}\label{eqn:embed}
\end{equation}

\subsection{Encoder Layer}
We use Bi-LSTM to encode our contexts and queries. For the contexts, first Bi-LSTM is mdoeling to capture the forward message of contexts, then set the final hidden state as $\vec{H}_{L}$, while do the same operation backward, then set the first hidden state as $\vec{H}_{1}$, finally we concatenate these two state as $\vec{H}_{1} \oplus {H}_{L}$. And queries likewise.

\begin{equation}
\begin{aligned}
\vec{E}_{Context} = \vec{H}_{1} \oplus \vec{H}_{L}\\
\vec{E}_{Query} = \vec{H}_{1} \oplus \vec{H}_{L}
\end{aligned}
\end{equation}

\section{Approach}

\begin{figure*}[t]
  \centering
  \includegraphics[scale=0.52]{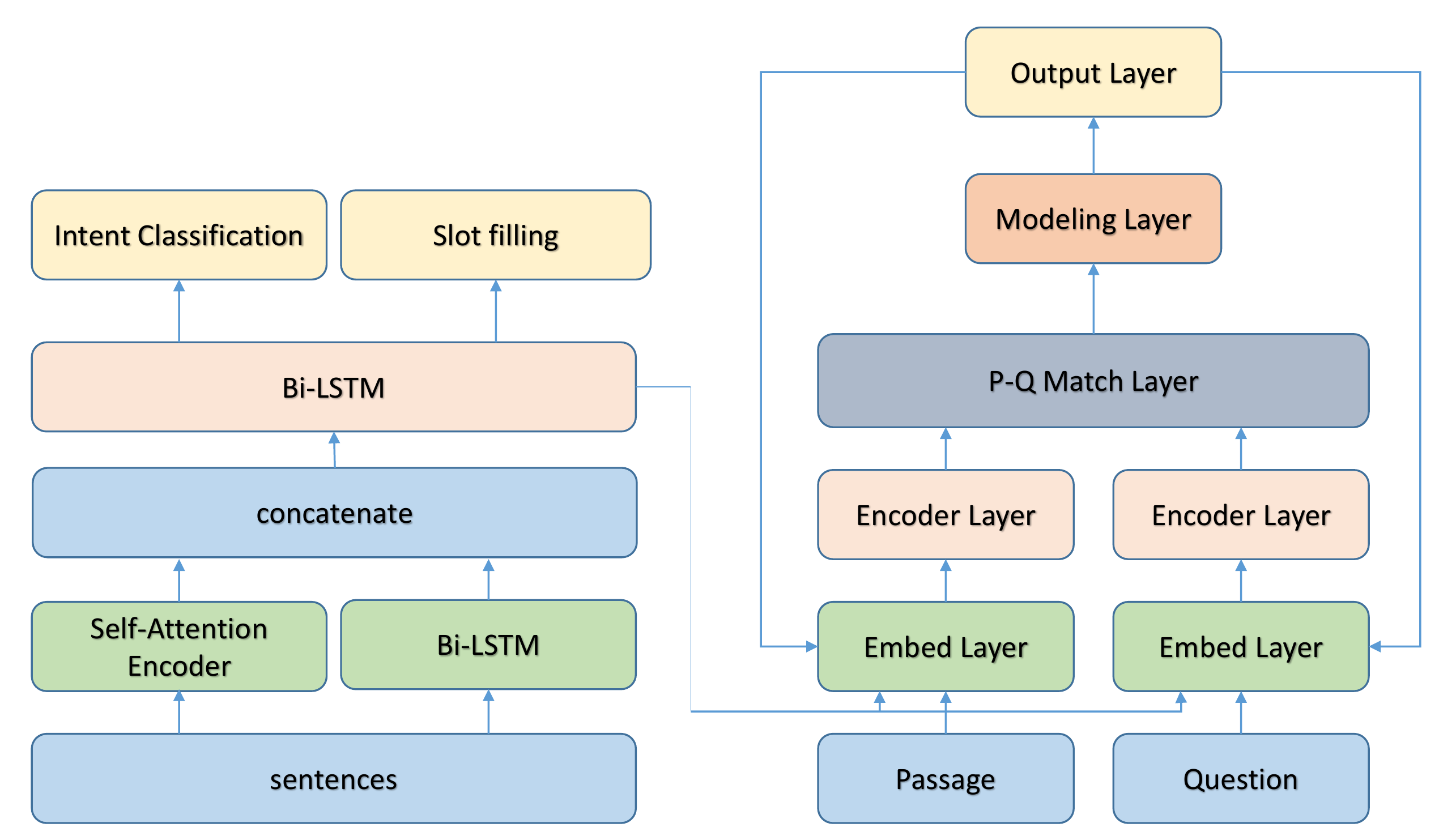}
  \caption{Our Machine Reading Comprehension Framework.}
  \label{fig:our_MRC}
\end{figure*}

From Figure~\ref{fig:our_MRC}, we can see that the proposed framework includes two modules, a basic Machine Reading Comprehension module and a jointly training of intent classification and slot filling module. In this section, we will illustrate the integration of the jointly training module to the MRC module.

\subsection{Encoder layer}
For the Strengthen-Encoder, intent detection task and slot filling task share the same encoder.
Following~\cite{Qin2019Stack-Propagation}, BiLSTM and self-attention are used for both advantages of temporal features and contextual information.

Given input sentence $\vec{X}=\left(\vec{y}_{1}, \vec{x}_{2}, \dots, \vec{x}_{T}\right) \in \R^{d \times T}$, BiLSTM~\cite{Hochreiter1997lstm} encodes it forwardly and backwardly to produce context-aware hidden state $\vec{H}=\left(\vec{h}_{1}, \vec{h}_{2}, \dots, \vec{h}_{T}\right) \in \R^{d \times T}$.

For self-attention mechanism~\cite{Vaswani2017attention}, we first map the input senquence $\vec{X} \in \R^{d \times T}$ to queries (Q), keys (K), values (V) vectors by using different linear projections, and the output $\vec{C} \in \R^{d \times T}$ is a weighted sum of V, the process is as follow:
\begin{equation}
\vec{C} = \text{softmax} \left(\frac{\vec{Q}\\ \vec{K}^{\top}}{\sqrt{d_{k}}}\right) \vec{V}
\end{equation}

After obtaining the output of self-attention and BiLSTM. We concatenate these two representations as the final encoding representation.

\subsection{Feature Integration Module}
We employ the model proposed by \citet{Qin2019Stack-Propagation}. As can be seen in Figure~\ref{fig:our_MRC}.
Specifically, Eqn \ref{eqn:embed} is used to embed the sentences, then we encode the sentences by self-attention layer and Bi-LSTM layer. After concatenation of encoder, they are transferred to different LSTM decoder, which represent different downstream tasks. 
When we train the MRC module, we inject the hidden state of SLU encoder to the MRC encoder, and the parameters of decoder layer in SLU module will not be updated.

\subsection{Training Objective}
After we inject the hidden state of encoder layer of SLU task to the embedding layer of MRC task, the loss function of our framework consists of three parts.
The total loss function is the sum of the negative log-likelihood loss of the three tasks:

\begin{equation}
\begin{split}
-\left(\sum_{\left(Y_{M}^{i}, X_{M}\right) \in \mathcal{M}} \log P\left(Y_{M}^{i} | X_{M}\right)\right.\\
\left.\quad+ \alpha \sum_{\left(Y_{d}, X_{d}\right) \in \mathcal{D}} \log P\left(Y_{d}^{*} | X_{d}\right)\right)
\end{split}
\end{equation}

where $\mathcal{M}$ is the set of MRC data and $\mathcal{D}$ is the set of SLU data, which includes intent classification data $\mathcal{I}$ and slot filling data $\mathcal{S}$.

\begin{table*}[t]
  \caption{Performance of our method and other models on dev set of SQuAD2.0}
  \label{experiment-table1}
  \centering
    \begin{tabular}{@{}l|c c@{}}
        \toprule
        Method   &  EM  & F1
        \\ \midrule [\heavyrulewidth]
        BiDAF \cite{Seo2017BiDAF}   & 57.60 &  61.10 \\
        w/ SLU & 59.80 (+2.2) & 62.80 (+1.7) \\  \midrule
        QA-Net \cite{Yu2018QA-Net} & 63.38 & 67.16  \\
        w/ SLU & 65.52 (+2.14) & 68.49 (+1.33)  \\  \midrule
        SAN \cite{Liu2017SAN} & 69.52 & 72.73  \\
        w/ SLU & 72.76 (+3.24) & 74.84 (+2.11)  \\  \midrule
        BERT \cite{Devlin2019BERT} & 80.01 & 83.06  \\
        w/ SLU & 81.90 (+1.89) & 84.68 (+1.62)  \\
       \bottomrule
    \end{tabular}
\end{table*}

\section{Experiment}

This section describes the experimental approach.
We explored the effectiveness of our encoder-strengthen framework on three machine reading comprehension models. The EM and F1 scores for these models are shown in table~\ref{experiment-table1}.  

\subsection{Datasets}
As a widely used MRC benchmark dataset, SQuAD 2.0~\cite{Rajpurkar2018squad2.0} combines the 100k questions from SQuAD 1.1~\cite{rajpurkar2016squad} with over 50k new.
Compared to SQuAD 1.1, SQuAD 2.0 requires models not only answer questions when possible, but also discards answers in passages that are not supported by answers.
We chose two metrics to evaluate the performance of the model: Exact Match (EM) and a F1 score.
An example of SQuAD2.0 is shown in Figure~\ref{fig:squad}.

We use SNIPS dataset~\cite{coucke2018snips} as the SLU dataset.
We follow the same format and partition as~\cite{goo2018slot}.
The dimension of the word embedding is 512 for SNIPS dataset.
The self-attentive encoder hidden units are set as 256.

\subsection{Experimental Details}
We train our Machine Reading Comprehension task together with Intent Classification task and Slot filling task. Specifically, we train MRC task in SQuAD 2.0 for one iteration, and both IC task and SF task for another. While training one certain task, the parameters of other tasks are fixed. Batch size is set to different according to different baseline systems. Training is terminated after $11$ epoch.

\subsection{Results}
\label{subsection1}

Experiment results are shown in Table~\ref{experiment-table1}. For the non pre-trained model, our framework has obvious improvement, especially, there are 3.24\% and 2.11\% absolute growth on EM and F1 for SAN model.
Besides, in order to prove the effectiveness of our framework, we also perform equivalent experiments on a strong pre-trained language model, i.e. BERT, which achieved great results in Machine Reading Comprehension and other tremendous NLP tasks.
From the result, we can see that though BERT has considerable results, its improvements are smaller compared to the other methods, in the assumption that BERT itself is one of an excellent sentence encoding model.

\section{Conclusion}
Due to the distantly loss back-propagating in reading comprehension, the encoder layer cannot learn effectively and directly supervised. Thus, the encoder layer can not learn the representation well at any time. 
In this paper, We propose a framework that can inject multi granularity message information to the encoding layer.
Empirical results show that our method can be effectively applied to existing MRC models.

\bibliography{compress}

\begin{thebibliography}{39}
\expandafter\ifx\csname natexlab\endcsname\relax\def\natexlab#1{#1}\fi

\bibitem[{Chen et~al.(2022{\natexlab{a}})Chen, Huang, Wu, Ge, and
  Zou}]{chen2022han}
D.~Chen, Z.~Huang, X.~Wu, S.~Ge, and Y.~Zou. 2022{\natexlab{a}}.
\newblock Towards joint intent detection and slot filling via higher-order
  attention.
\newblock In \emph{Proc. of IJCAI}.

\bibitem[{Chen et~al.(2022{\natexlab{b}})Chen, Huang, and
  Zou}]{chen2022bilinear}
D.~Chen, Z.~Huang, and Y.~Zou. 2022{\natexlab{b}}.
\newblock Leveraging bilinear attention to improve spoken language
  understanding.
\newblock In \emph{Proc. of ICASSP}.

\bibitem[{Chen et~al.(2022{\natexlab{c}})Chen, Zhou, and Zou}]{chen2022joint}
L.~Chen, P.~Zhou, and Y.~Zou. 2022{\natexlab{c}}.
\newblock Joint multiple intent detection and slot filling via
  self-distillation.
\newblock In \emph{Proc. of ICASSP}.

\bibitem[{Chen et~al.(2021)Chen, Liu, You, Zhou, and Zou}]{chen2021adaptive}
N.~Chen, F.~Liu, C.~You, P.~Zhou, and Y.~Zou. 2021.
\newblock Adaptive bi-directional attention: Exploring multi-granularity
  representations for machine reading comprehension.
\newblock In \emph{Proc. of ICASSP}.

\bibitem[{Chen et~al.(2019)Chen, Zhuo, and Wang}]{Chen2019jointbert}
Q.~Chen, Z.~Zhuo, and W.~Wang. 2019.
\newblock {BERT} for joint intent classification and slot filling.
\newblock \emph{arXiv preprint arXiv:1902.10909}.

\bibitem[{Chong et~al.(2022)Chong, Wang, Zhou, and Zeng}]{chong2022masked}
D.~Chong, H.~Wang, P.~Zhou, and Q.~Zeng. 2022.
\newblock Masked spectrogram prediction for self-supervised audio pre-training.
\newblock \emph{arXiv preprint arXiv:2204.12768}.

\bibitem[{Coucke et~al.(2018)Coucke, Saade, Ball, Bluche, Caulier, Leroy,
  Doumouro, Gisselbrecht, Caltagirone, Lavril et~al.}]{coucke2018snips}
A.~Coucke, A.~Saade, A.~Ball, T.~Bluche, A.~Caulier, D.~Leroy, C.~Doumouro,
  T.~Gisselbrecht, F.~Caltagirone, T.~Lavril, et~al. 2018.
\newblock Snips voice platform: an embedded spoken language understanding
  system for private-by-design voice interfaces.
\newblock \emph{arXiv preprint arXiv:1805.10190}.

\bibitem[{Devlin et~al.(2019)Devlin, Chang, Lee, and
  Toutanova}]{Devlin2019BERT}
J.~Devlin, M.~Chang, K.~Lee, and K.~Toutanova. 2019.
\newblock {BERT:} pre-training of deep bidirectional transformers for language
  understanding.
\newblock In \emph{Proc. of NAACL}.

\bibitem[{Goo et~al.(2018)Goo, Gao, Hsu, Huo, Chen, Hsu, and
  Chen}]{goo2018slot}
C.~Goo, G.~Gao, Y.~Hsu, C.~Huo, T.~Chen, K.~Hsu, and Y.~Chen. 2018.
\newblock Slot-gated modeling for joint slot filling and intent prediction.
\newblock In \emph{Proc. of NAACL}.

\bibitem[{Hochreiter and Schmidhuber(1997)}]{Hochreiter1997lstm}
S.~Hochreiter and J.~Schmidhuber. 1997.
\newblock Long short-term memory.
\newblock \emph{Neural Computation}.

\bibitem[{Hou et~al.(2020)Hou, Huang, Shang, Jiang, Chen, and
  Liu}]{hou2020dynabert}
L.~Hou, Z.~Huang, L.~Shang, X.~Jiang, X.~Chen, and Q.~Liu. 2020.
\newblock Dynabert: Dynamic {BERT} with adaptive width and depth.
\newblock In \emph{Proc. of NeurIPS}.

\bibitem[{Huang et~al.(2021{\natexlab{a}})Huang, Hou, Shang, Jiang, Chen, and
  Liu}]{huang2021ghostbert}
Z.~Huang, L.~Hou, L.~Shang, X.~Jiang, X.~Chen, and Q.~Liu. 2021{\natexlab{a}}.
\newblock Ghostbert: Generate more features with cheap operations for {BERT}.
\newblock In \emph{Proc. of ACL}.

\bibitem[{Huang et~al.(2021{\natexlab{b}})Huang, Liu, Wu, Ge, Wang, Fan, and
  Zou}]{huang2021audioMRC2}
Z.~Huang, F.~Liu, X.~Wu, S.~Ge, H.~Wang, W.~Fan, and Y.~Zou.
  2021{\natexlab{b}}.
\newblock Audio-oriented multimodal machine comprehension: Task, dataset and
  model.
\newblock \emph{arXiv preprint arXiv:2107.01571}.

\bibitem[{Huang et~al.(2021{\natexlab{c}})Huang, Liu, Wu, Ge, Wang, Fan, and
  Zou}]{huang2021audioMRC}
Z.~Huang, F.~Liu, X.~Wu, S.~Ge, H.~Wang, W.~Fan, and Y.~Zou.
  2021{\natexlab{c}}.
\newblock Audio-oriented multimodal machine comprehension via dynamic inter-
  and intra-modality attention.
\newblock In \emph{Proc. of AAAI}.

\bibitem[{Huang et~al.(2021{\natexlab{d}})Huang, Liu, Zhou, and
  Zou}]{huang2021sentiment}
Z.~Huang, F.~Liu, P.~Zhou, and Y.~Zou. 2021{\natexlab{d}}.
\newblock Sentiment injected iteratively co-interactive network for spoken
  language understanding.
\newblock In \emph{Proc. of ICASSP}.

\bibitem[{Huang et~al.(2020)Huang, Liu, and Zou}]{huang2020FLSLU}
Z.~Huang, F.~Liu, and Y.~Zou. 2020.
\newblock Federated learning for spoken language understanding.
\newblock In \emph{Proc. of COLING}.

\bibitem[{Huang et~al.(2022)Huang, Rao, Raju, Zhang, Bui, and
  Lee}]{huang2022slt}
Z.~Huang, M.~Rao, A.~Raju, Z.~Zhang, B.~Bui, and C.~Lee. 2022.
\newblock {MTL-SLT:} multi-task learning for spoken language tasks.
\newblock In \emph{Proc. of ACL}.

\bibitem[{Kadlec et~al.(2016)Kadlec, Schmid, Bajgar, and
  Kleindienst}]{kadlec2016text}
R.~Kadlec, M.~Schmid, O.~Bajgar, and J.~Kleindienst. 2016.
\newblock Text understanding with the attention sum reader network.
\newblock \emph{arXiv preprint arXiv:1603.01547}.

\bibitem[{Liu et~al.(2017)Liu, Shen, Duh, and Gao}]{Liu2017SAN}
X.~Liu, Y.~Shen, K.~Duh, and J.~Gao. 2017.
\newblock Stochastic answer networks for machine reading comprehension.
\newblock \emph{CoRR}.

\bibitem[{Qin et~al.(2019)Qin, Che, Li, Wen, and
  Liu}]{Qin2019Stack-Propagation}
L.~Qin, W.~Che, Y.~Li, H.~Wen, and T.~Liu. 2019.
\newblock A stack-propagation framework with token-level intent detection for
  spoken language understanding.
\newblock \emph{CoRR}.

\bibitem[{Rajpurkar et~al.(2018)Rajpurkar, Jia, and
  Liang}]{Rajpurkar2018squad2.0}
P.~Rajpurkar, R.~Jia, and P.~Liang. 2018.
\newblock Know what you don't know: Unanswerable questions for squad.
\newblock In \emph{ACL}.

\bibitem[{Rajpurkar et~al.(2016)Rajpurkar, Zhang, Lopyrev, and
  Liang}]{rajpurkar2016squad}
P.~Rajpurkar, J.~Zhang, K.~Lopyrev, and P.~Liang. 2016.
\newblock Squad: 100,000+ questions for machine comprehension of text.
\newblock \emph{arXiv preprint arXiv:1606.05250}.

\bibitem[{Seo et~al.(2017)Seo, Kembhavi, Farhadi, and
  Hajishirzi}]{Seo2017BiDAF}
M.~Seo, A.~Kembhavi, A.~Farhadi, and H.~Hajishirzi. 2017.
\newblock Bidirectional attention flow for machine comprehension.
\newblock In \emph{ICLR}.

\bibitem[{Sordoni et~al.(2016)Sordoni, Bachman, Trischler, and
  Bengio}]{sordoni2016iterative}
A.~Sordoni, P.~Bachman, A.~Trischler, and Y.~Bengio. 2016.
\newblock Iterative alternating neural attention for machine reading.
\newblock \emph{arXiv preprint arXiv:1606.02245}.

\bibitem[{Vaswani et~al.(2017)Vaswani, Shazeer, Parmar, Uszkoreit, Jones,
  Gomez, Kaiser, and Polosukhin}]{Vaswani2017attention}
A.~Vaswani, N.~Shazeer, N.~Parmar, J.~Uszkoreit, L.~Jones, A.~Gomez, L.~Kaiser,
  and I.~Polosukhin. 2017.
\newblock Attention is all you need.
\newblock In \emph{Proc. of NeurIPS}.

\bibitem[{Wang and Jiang(2017)}]{Wang2017MatchLSTM}
S.~Wang and J.~Jiang. 2017.
\newblock Machine comprehension using match-lstm and answer pointer.
\newblock In \emph{ICLR}.

\bibitem[{Wang et~al.(2017)Wang, Yang, Wei, Chang, and Zhou}]{Wang2017R-Net}
W.~Wang, N.~Yang, F.~Wei, B.~Chang, and M.~Zhou. 2017.
\newblock Gated self-matching networks for reading comprehension and question
  answering.
\newblock In \emph{ACL}.

\bibitem[{Xia et~al.(2019)Xia, Li, and Zhang}]{xia2019SRL}
Q.~Xia, Z.~Li, and M.~Zhang. 2019.
\newblock A syntax-aware multi-task learning framework for chinese semantic
  role labeling.
\newblock In \emph{Proc. of EMNLP}.

\bibitem[{Xie et~al.(2022)Xie, Zhou, and Kim}]{xie2022decoupled}
Y.~Xie, P.~Zhou, and S.~Kim. 2022.
\newblock Decoupled side information fusion for sequential recommendation.
\newblock In \emph{Proc. of SIGIR}.

\bibitem[{Xiong et~al.(2017)Xiong, Zhong, and Socher}]{Xiong2017dynamic}
C.~Xiong, V.~Zhong, and R.~Socher. 2017.
\newblock Dynamic coattention networks for question answering.
\newblock In \emph{Proc. of ICLR}.

\bibitem[{Xu et~al.(2021)Xu, Zhou, You, and Zou}]{xu2021semantic}
W.~Xu, P.~Zhou, C.~You, and Y.~Zou. 2021.
\newblock Semantic transportation prototypical network for few-shot intent
  detection.
\newblock In \emph{Proc. of Interspeech}.

\bibitem[{Yang et~al.(2019)Yang, Dai, Yang, Carbonell, Salakhutdinov, and
  Le}]{yang2019xlnet}
Z.~Yang, Z.~Dai, Y.~Yang, J.~Carbonell, R.~Salakhutdinov, and Q.~Le. 2019.
\newblock Xlnet: Generalized autoregressive pretraining for language
  understanding.
\newblock In \emph{Proc. of NeurIPS}.

\bibitem[{Yu et~al.(2018)Yu, Dohan, Luong, Zhao, Chen, Norouzi, and
  Le}]{Yu2018QA-Net}
A.~Yu, D.~Dohan, M.~Luong, R.~Zhao, K.~Chen, M.~Norouzi, and Q.~Le. 2018.
\newblock Qanet: Combining local convolution with global self-attention for
  reading comprehension.
\newblock In \emph{ICLR}.

\bibitem[{Zeng et~al.(2022)Zeng, Chong, Zhou, and Yang}]{zeng2022low}
Q.~Zeng, D.~Chong, P.~Zhou, and J.~Yang. 2022.
\newblock Low-resource accent classification in geographically-proximate
  settings: A forensic and sociophonetics perspective.
\newblock \emph{Interspeech}.

\bibitem[{Zhou et~al.(2022{\natexlab{a}})Zhou, Chong, Wang, and
  Zeng}]{zhou2022calibrate}
P.~Zhou, D.~Chong, H.~Wang, and Q.~Zeng. 2022{\natexlab{a}}.
\newblock Calibrate and refine! a novel and agile framework for asr-error
  robust intent detection.
\newblock In \emph{Proc. of Interspeech}.

\bibitem[{Zhou et~al.(2022{\natexlab{b}})Zhou, Gao, Xie, Ye, Hua, and
  Kim}]{zhou2022equivariant}
P.~Zhou, J.~Gao, Y.~Xie, Q.~Ye, Y.~Hua, and S.~Kim. 2022{\natexlab{b}}.
\newblock Equivariant contrastive learning for sequential recommendation.
\newblock \emph{arXiv preprint arXiv:2211.05290}.

\bibitem[{Zhou et~al.(2020)Zhou, Huang, Liu, and Zou}]{zhou2020pin}
P.~Zhou, Z.~Huang, F.~Liu, and Y.~Zou. 2020.
\newblock {PIN:} {A} novel parallel interactive network for spoken language
  understanding.
\newblock In \emph{Proc. of ICPR}.

\bibitem[{Zhou et~al.(2022{\natexlab{c}})Zhou, Wang, Chong, Guo, Hua, Su, Teng,
  Wu, and Yang}]{zhou2022mets}
P.~Zhou, Z.~Wang, D.~Chong, Z.~Guo, Y.~Hua, Z.~Su, Z.~Teng, J.~Wu, and J.~Yang.
  2022{\natexlab{c}}.
\newblock Mets-cov: A dataset of medical entity and targeted sentiment on
  covid-19 related tweets.
\newblock \emph{NeurIPS}.

\bibitem[{Zhu et~al.(2022)Zhu, Xu, Cheng, Song, and Zou}]{Zhu2022graph}
Z.~Zhu, W.~Xu, X.~Cheng, T.~Song, and Y.~Zou. 2022.
\newblock A dynamic graph interactive framework with label-semantic injection
  for spoken language understanding.
\newblock \emph{arXiv preprint arXiv:2211.04023}.

\end{thebibliography}
\bibliographystyle{acl_natbib}

\end{document}